\documentclass{article}
\usepackage{ijcai23}

\usepackage{times}
\usepackage{soul}
\usepackage{url}
\usepackage[hidelinks]{hyperref}
\usepackage[utf8]{inputenc}
\usepackage[small]{caption}
\usepackage{graphicx}
\usepackage{amsfonts}
\usepackage{amsmath}
\usepackage{amsthm}
\usepackage{booktabs}
\usepackage{algorithm}
\usepackage{algorithmic}
\usepackage[switch]{lineno}
\usepackage{enumitem}

\usepackage[edges]{forest}
\definecolor{hiddendraw}{RGB}{205, 44, 36}
\usepackage{color}
\usepackage{multirow}
\usepackage{subfigure}

\newcommand{\citet}[1]{\citeauthor{#1}~\shortcite{#1}}
\newcommand{\expect}{\ensuremath{\mathbb{E}}}

\tikzset{%
    parent/.style =          {align=center,rounded corners=3pt, line width=0.3mm, fill=gray!10,draw=gray!80},
    child/.style =           {align=center,rounded corners=3pt, fill=blue!10,draw=blue!80,line width=0.3mm},
    grandchild/.style =      {align=center,text width=2cm,rounded corners=3pt},
    greatgrandchild/.style = {align=center,text width=1.5cm,rounded corners=3pt},
    greatgrandchild2/.style = {align=center,text width=1.5cm,rounded corners=3pt},    
    referenceblock/.style =  {align=center,text width=1.5cm,rounded corners=2pt},
    pretrain/.style =           {align=center,rounded corners=3pt, fill=blue!10,draw=blue!80,line width=0.2mm},   
    pretrain_work/.style =           {align=center,rounded corners=3pt, fill=blue!10,draw=blue!0,line width=0.2mm},  
    template/.style =   {align=center,rounded corners=3pt, fill=red!10,draw=red!80,line width=0.2mm},   
    template_work/.style =  {align=center,rounded corners=3pt, fill=red!10,draw=red!0,line width=0.2mm},    
    answer/.style =           {align=center,rounded corners=3pt, fill= cyan!10,draw= cyan!80,line width=0.2mm},   
    answer_work/.style =           {align=center,rounded corners=3pt, fill= cyan!10,draw= cyan!0,line width=0.2mm},      
    multiple/.style =           {align=center,rounded corners=3pt, fill= orange!10,draw= orange!80,line width=0.2mm},   
    multiple_work/.style =           {align=center,rounded corners=3pt, fill= orange!10,draw= orange!0,line width=0.2mm},        
    tuning/.style =           {align=center,rounded corners=3pt, fill= magenta!10,draw= magenta!80,line width=0.2mm},   
    tuning_work/.style =           {align=center,rounded corners=3pt, fill= magenta!10,draw= magenta!0,line width=0.2mm},          
}

\title{Recent Advances in Direct Speech-to-text Translation}

\author{
    Chen Xu$^1$\footnote{Equal Contribution. The order is decided by the dice roll.}\and
    Rong Ye$^{2}$\footnotemark[1]\and
    Qianqian Dong$^2$\footnotemark[1]\and
    Chengqi Zhao$^2$\and
    Tom Ko$^2$\and \\
    Mingxuan Wang$^2$\footnote{Corresponding Authors.}\and
    Tong Xiao$^{1,3}$\footnotemark[2] \textnormal{and}
    Jingbo Zhu$^{1,3}$
    \affiliations
    $^1$School of Computer Science and Engineering, Northeastern University, Shenyang, China\\
    $^2$ByteDance AI Lab \\
    $^3$NiuTrans Research, Shenyang, China \\
    \emails
    xuchennlp@outlook.com, 
    \{xiaotong, zhujingbo\}@mail.neu.edu.cn \\
    \{yerong, dongqianqian, zhaochengqi.d, tom.ko, wangmingxuan.89\}@bytedance.com\\
}

\begin{document}

\maketitle

\begin{abstract}
Recently, speech-to-text translation has attracted more and more attention and many studies have emerged rapidly. In this paper, we present a comprehensive survey on direct speech translation aiming to summarize the current state-of-the-art techniques. First, we categorize the existing research work into three directions based on the main challenges --- modeling burden, data scarcity, and application issues. To tackle the problem of modeling burden, two main structures have been proposed, encoder-decoder framework (Transformer and the variants) and multitask frameworks. For the challenge of data scarcity, recent work resorts to many sophisticated techniques, such as data augmentation, pre-training, knowledge distillation, and multilingual modeling. We analyze and summarize the application issues, which include real-time, segmentation, named entity, gender bias, and code-switching. Finally, we discuss some promising directions for future work.
\end{abstract}

\section{Introduction}

Speech-to-text translation (ST) is a task that aims to translate speech in one language to text in another language.
It has numerous practical applications, including global communication, language learning, and accessibility for non-native speakers. 

Early solutions for speech translation are to break down the task into smaller and more manageable sub-tasks, such as automatic speech recognition (ASR) and machine translation (MT). 
This is the idea of the cascaded system.
For example, to fulfill the ST task, we can cascade an ASR system to transcribe speech into text with an MT system to translate text into another language in tandem~\cite{stentiford1988machine}.
Research on cascaded systems mainly aims at solving the problem of error accumulation, such as utilizing multiple recognition results and training robust MT models.


Meanwhile, the end-to-end speech translation~(E2E ST) model, which eliminates the need for intermediate steps~(\textit{e.g.}, ASR and MT), is designed and also has the potential to eliminate error accumulation.
In addition, it also has the advantage of reduced latency, more contextual modeling~\cite{Bentivogli_ACL2021}, and applicability to unwritten languages~\cite{Berard_arxiv2016}.
In recent years, research on end-to-end models in speech translation has gained momentum, leading to diversity in model architectures and training methods. 
However, a comprehensive survey that thoroughly reviews their motivations and practices is currently lacking.

\noindent\paragraph{Basic modeling.}
ST corpus usually contains the source speech $\mathbf{s}$, transcription $\mathbf{x}$, and translation $\mathbf{y}$. 
The basic model framework of E2E ST is mainly based on the encoder-decoder structure.
The encoder encodes the speech input into a sequence of hidden states, and the decoder outputs the final translation result condition on the hidden states, which is basically autoregressive.
The objective training function of the ST model $\theta$ is the negative log-likelihood loss:
\begin{equation*}
    L_{\theta} =  - \expect_{s,y} \log p(\mathbf{y}|\mathbf{s};\theta) =  - \expect_{s,y} \sum_{t=1}^T \log p(y_t|\mathbf{y_{<t}}, \mathbf{s}; \theta)
\end{equation*}
where $T$ is the length of $\mathbf{y}$. In the inference stage, we usually apply beam search to generate target sentences.

\begin{figure*}[th]
  \centering
  \resizebox{0.9\textwidth}{!}{
\begin{forest}
  forked edges,
  for tree={
    grow=east,
    reversed=true, 
    anchor=base west,
    parent anchor=east,
    child anchor=west,
    base=left,
    font=\footnotesize,
    rectangle,
    draw=hiddendraw,
    rounded corners, 
    align=left,
    minimum width=2.5em,
    minimum height=1.2em,
    s sep=6pt,
    inner xsep=3pt,
    inner ysep=1.5pt,
  },
  where level=1{text width=10.3em}{},
  where level=2{text width=11em}{},
  where level=3{text width=10.6em}{},
  where level=4{text width=6.7em}{},
  where level=5{font=\scriptsize}{},
  [\textbf{Approaches}, fill=gray!45, parent
    [Tackling \textbf{Modeling Burden}, for tree={fill=red!45,template}
      [Transformer and Variants
        [Speech-Transformer]
        [Conformer]
        [SSL-Transformer]
      ]
      [Multitask Frameworks
        [Decoupled Decoder
            [Two-pass Decoder]
            [Dual Decoder]
          ]
        [Decoupled Encoder]
        [Two-stream Encoder
            [Speech Encoder]
            [Text Encoder]
            [Interaction]
        ]
       ]
       [Non-autoregressive Modeling]
    ]
    [Tackling \textbf{Data Scarcity},  for tree={multiple}
        [Data Augmentation
            [Expanding ST Data]
            [Speech Augmentation]
        ]
        [Pre-training
            [Separate Pre-training]
            [Joint Pre-training]
        ]
        [Knowledge Distillation]
        [Multilingual Training]
    ]
    [Tackling \textbf{Application Issues}, for tree={answer}
     [
     Real-time
     ]
      [Segmentation
      ]
      [Named Entity
      ]
      [Gender Bias]
      [Code-Switching]
     ]
    ]
\end{forest}
}
\caption{Taxonomy of speech-to-text translation approaches.}
\label{fig:taxonomy_of_training_design2}
\end{figure*}
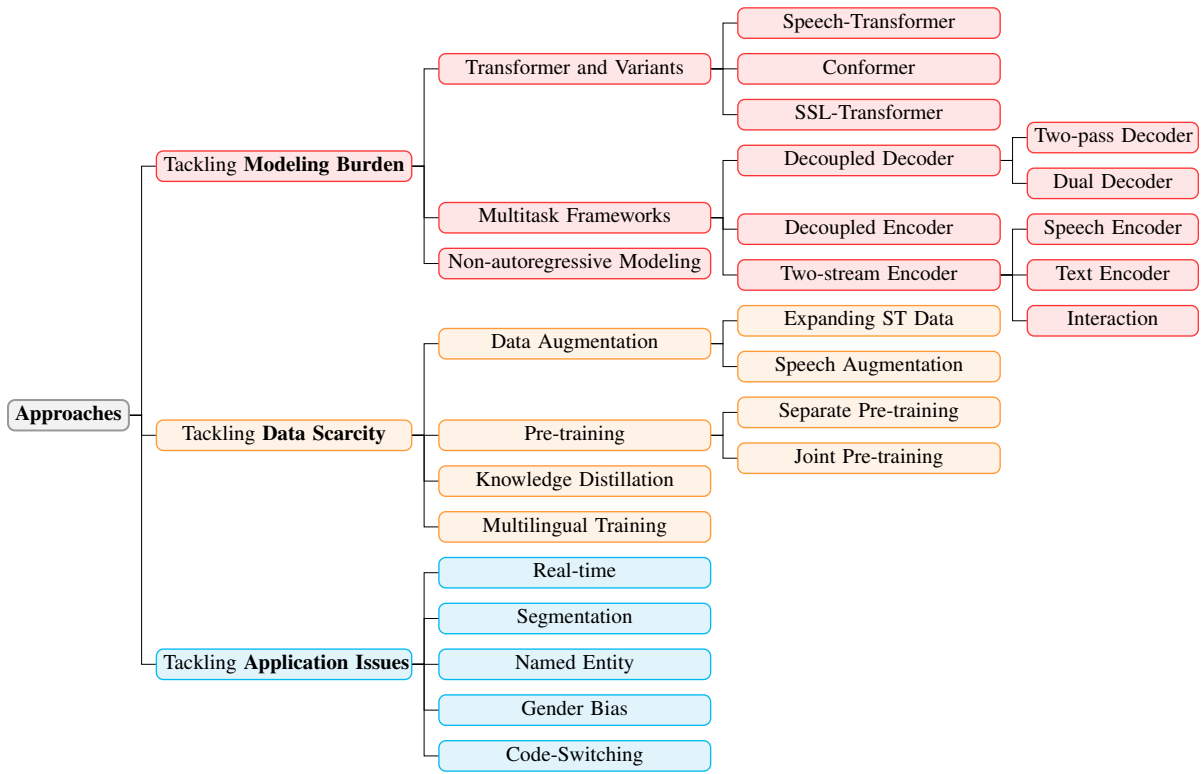

However, we find that training an E2E ST model is not easy.
Although the study also confirms that the performance of the end-to-end model is approaching the results of the cascaded solution, it is still not the best-performing technology~\cite{Anastasopoulos_IWSLT2022}.
Existing literature mainly attributes and attempts to address the following challenges:

\begin{itemize}
    \item \textbf{Modeling burden}:
        Conventional cascaded systems decouple it into ASR and MT models, while it is non-trivial and burdensome for speech translation in an E2E manner.
        This is because the E2E model requires both cross-modal and cross-lingual mapping at the same time.
        Training the E2E model often encounters poor convergence and low performance~\cite{Berard_arxiv2016,Weiss_ISCA2017}.
    \item \textbf{Data scarcity}:
        Annotating speech translation data is demanding, so labeled parallel data for training is scarce.         
        For example, an ASR dataset like Librispeech\footnote{https://www.openslr.org/12} contains 960 hours of speech~\cite{Panayotov_ICASSP2015} and the MT dataset typically has millions of bi-texts, while the ST dataset, such as MuST-C~\cite{Gangi_NAACL2019} contains only about 400 hours of speech with 230k utterances\footnote{https://ict.fbk.eu/must-c/}.
        Data scarcity results in the E2E model being inferior to cascaded systems trained on abundant ASR and MT data \cite{sperber2020speech}, which is more severe in the industry.

    \item \textbf{Application issues}:
        In addition to performance, there are other considerations in the practical implementation, such as real-time, long-form audio segmentation, gender bias, named entity translation, and code-switching speech. 
\end{itemize}

Correspondingly, in this paper, we provide a comprehensive survey of how previous work has tackled the above challenges, and hope to suggest some directions for future research in this community.
As shown in the taxonomy in Figure~\ref{fig:taxonomy_of_training_design2}, our survey is developed as follows.

\begin{itemize}[itemsep=1pt, leftmargin=12pt, parsep=0pt, topsep=1pt]
    \item Section~\ref{sec:model} describes how to alleviate the modeling burden challenge in the existing literature. Modeling methods can be divided into three categories: Transformer and the variants, multitask frameworks, and non-autoregressive modeling.
    \item Section~\ref{sec:training} summarizes approaches to tackle the problem of data scarcity, including data augmentation, pre-training, knowledge distillation, and multilingual training.
    \item Section~\ref{sec:application} briefly introduces application issues (real-time, segmentation, named entity, gender bias, code-switching) in practice and recent related work. 
    \item Section~\ref{sec:future} anticipates some promising directions for future ST research.
\end{itemize}

\section{Tackling Modeling Burden}
\label{sec:model}

For the long sequence input, like speech, we require high-capacity E2E models, typically Transformer-based ST models and their variants~(Section~\ref{sec:model:enc_dec}).
Additionally, to address the issue of modeling burden, the existing literature generally employs a multitask framework to make modifications to the original Transformer-based model. We categorize and introduce them in Section~\ref{sec:model:mlt}.
Finally, for the consideration of decoding efficiency, there are also non-autoregressive models introduced in Section~\ref{sec:model:nat}.

\subsection{Transformer and Variants}
\label{sec:model:enc_dec}

The ST task is based on the sequence-to-sequence modeling, which typically adopts the encoder-decoder architecture, as shown in Figure~\ref{fig:enc_dec}.
Among many well-established networks, Transformer \cite{Vaswani_nips2017} is chosen for its state-of-the-art performance across almost all sequence generation tasks.
Several variants have been proposed to make Transformer more suitable for speech modeling.
Here we highlight Speech-Transformer, Conformer, and SSL-Transformer.

\paragraph{Speech-Transformer.}~
Speech-Transformer~\cite{di2019adapting} is built on top of the text-to-text Transformer~\cite{Vaswani_nips2017}. 
The major difference is that the sequence of audio features (\textit{e.g.} Fbank) is first compressed by the convolutional layers (typically two layers with a stride of 2, compressing the length by a factor of 4) and a normalization layer before the self-attention encoder.
   
\paragraph{Conformer.}~
Conformer is a convolution-augmented Transformer model~\cite{Gulati_ISCA2020}. 
The main feature of the Conformer is the convolution module, which is inserted between the multi-head self-attention module and the feed-forward layer of each encoder block. 
The convolution module has the attention and convolution modules,  sandwiched by two Macaron-net style feed-forward layers and the residual connections.
The combination of CNN and Transformer helps to model both local and global information, which is suitable for encoding long-sequence speech.


\paragraph{SSL-Transformer.}~
With the success of self-supervised learned (SSL) speech representations, such as Wav2vec-family~\cite{Steffen_Interspeech2019,baevski2020wav2vec} and HuBERT~\cite{hsu2021hubert} on ASR, they have also been utilized in the encoder of ST models, which we collectively refer to as \textit{SSL-Transformer}.
The original audio waveform is fed into the SSL model, which subsequently processes the waveform through several convolutional layers and Transformer encoder layers to extract speech features.
In the SSL-Transformer model, the SSL model can be incorporated into the decoder either as a standalone encoder~\cite{Wu_ISCA2020,Nguyen_ISCA2020,wang2021large} or as a speech feature extractor, which is then connected to the whole Transformer~\cite{han2021learning,Ye_interspeech2021}.

\subsection{Multitask Frameworks}
\label{sec:model:mlt}


The idea of the multitask framework is to utilize related auxiliary tasks to enhance the target task. For ST, the auxiliary tasks are often ASR and MT. 
As for the model structure, some parameters of the target and auxiliary task modules can be shared, while there are parts of the modules that remain independent. 
We summarize the multitasking frameworks in the extant literature~(in chronological order), which can be broadly classified into the following three types, namely \textbf{decoupled decoder}~(Figure~\ref{fig:decouple_dec}), \textbf{decoupled encoder}~(Figure~\ref{fig:decouple_enc}), and \textbf{two-stream encoder}~(Figure~\ref{fig:2_enc}).

\subsubsection{Decoupled Decoder}
\label{sec:model:mlt:decouple_dec}

\begin{figure}
    \centering
    \subfigure[Encoder-Decoder]{
        \begin{minipage}[t]{0.45\linewidth}
            \centering
            \includegraphics[scale=0.27]{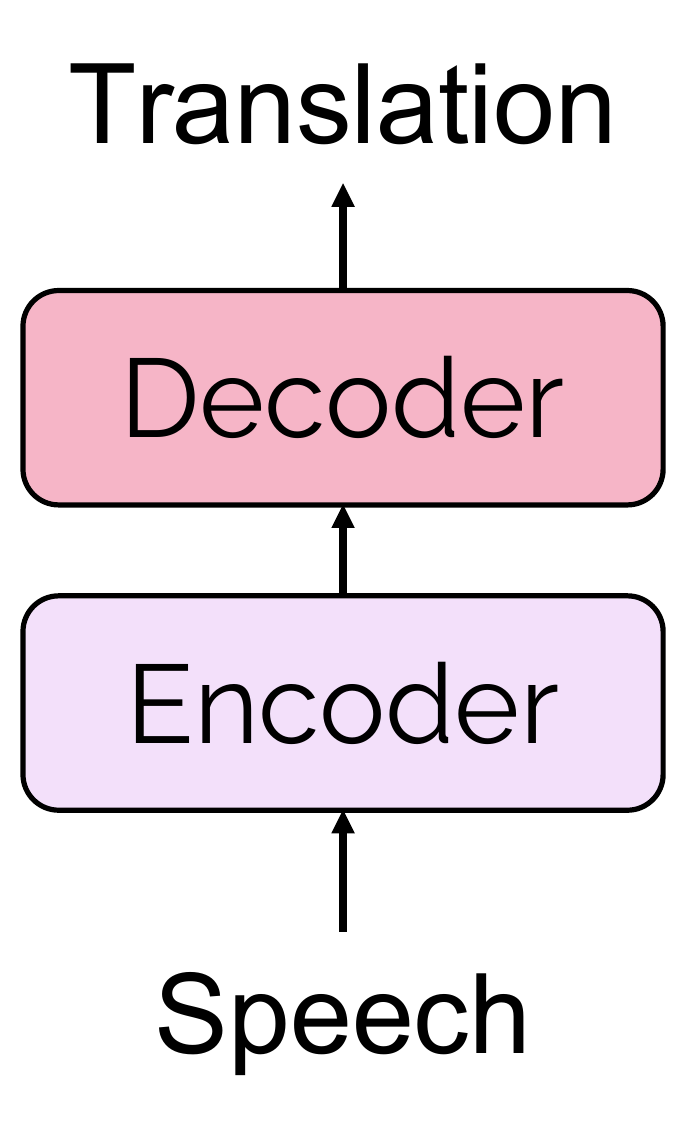}
        \end{minipage}
        \label{fig:enc_dec}
    }
    \subfigure[Decoupled Decoder]{
        \begin{minipage}[t]{0.45\linewidth}
            \centering
            \includegraphics[scale=0.25]{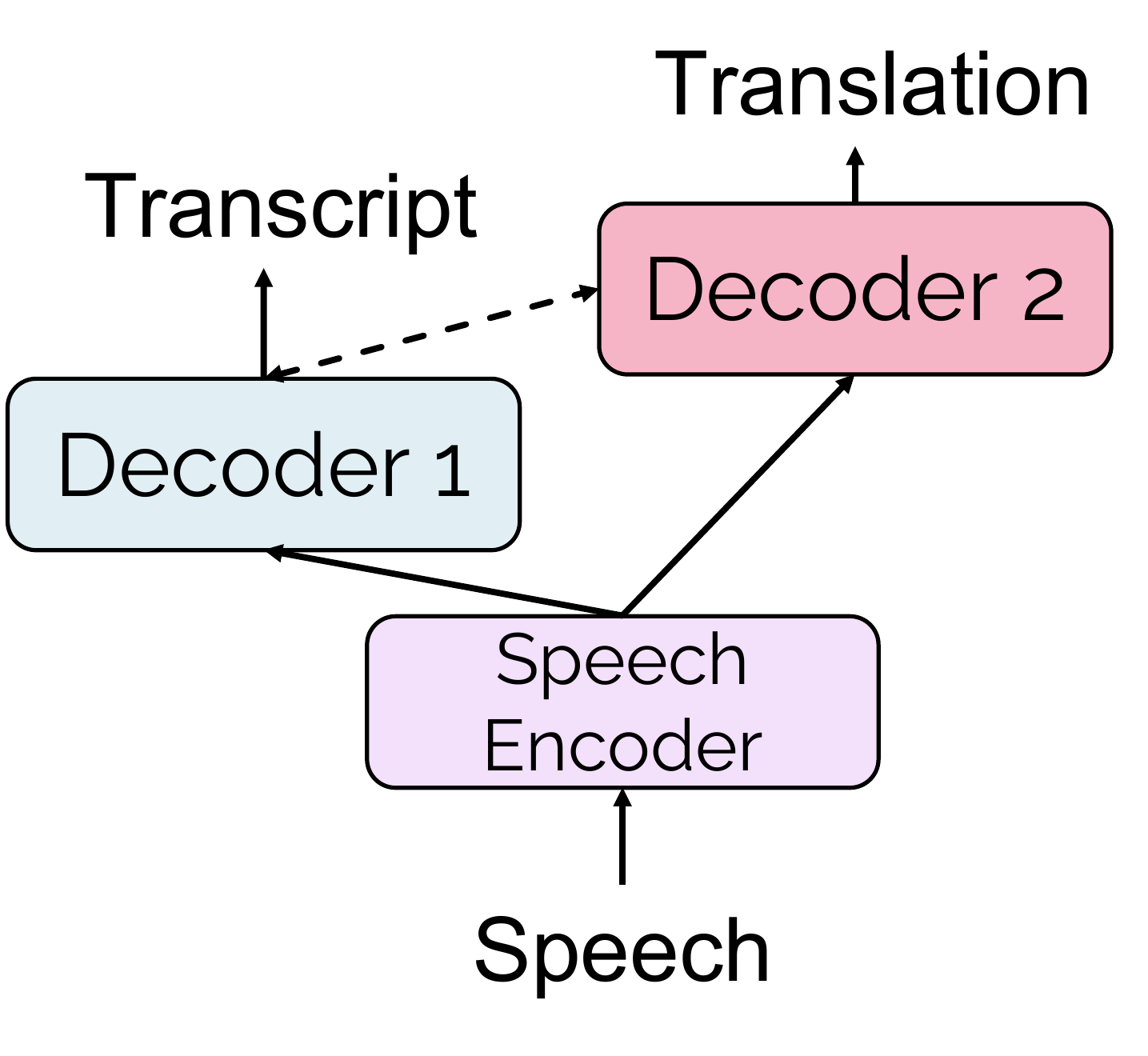}
        \end{minipage}
        \label{fig:decouple_dec}
    }

    \subfigure[Decoupled Encoder]{
        \begin{minipage}[t]{0.45\linewidth}
            \includegraphics[scale=0.25]{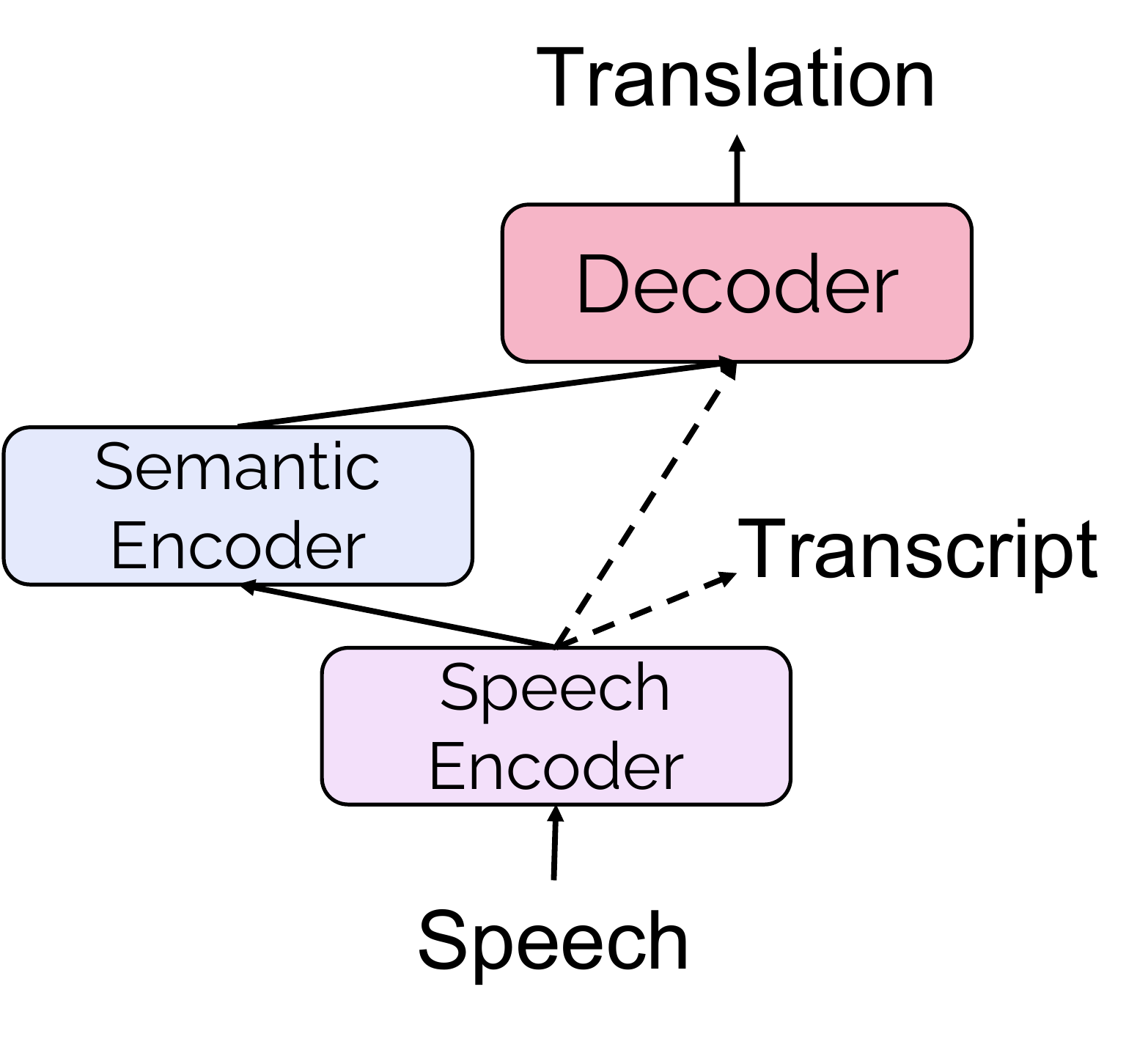}
        \end{minipage}
        \label{fig:decouple_enc}
    }
    \subfigure[Two-stream Encoder]{
        \begin{minipage}[t]{0.45\linewidth}
            \centering
            \includegraphics[scale=0.25]{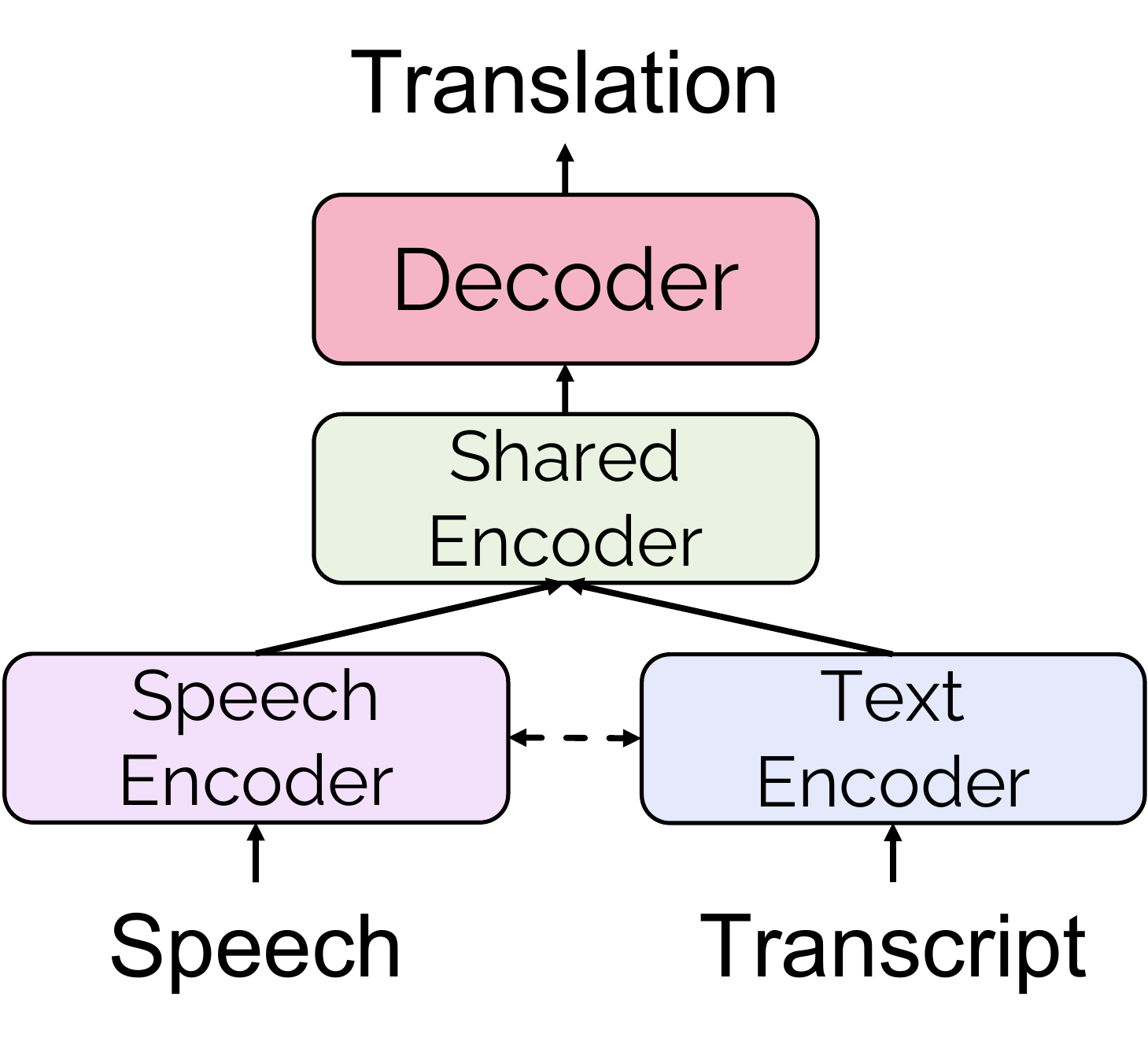}
        \end{minipage}
        \label{fig:2_enc}
    }
    \caption{Schematic diagram of the model architectures. The dashed lines indicate possible interactions between the modules.}
    \label{fig:all_model_structures}
\end{figure}

To facilitate the burden of direct cross-modal and cross-lingual modeling, the additional decoder is introduced to guide the learning of transcript, while the model is still trained in an E2E manner, as shown in Figure~\ref{fig:decouple_dec}.
The naive approach only decodes the transcription as the auxiliary task~\cite{Weiss_ISCA2017}, and then the researchers further explore how to better prompt translation by the generated transcription (\textit{two-pass decoder})~\cite{Anastasopoulos_NAACL2018} or generate transcription and translation synchronously (\textit{dual decoder})~\cite{Liu_AAAI2020}.

\begin{itemize}
    \item \textbf{Two-pass decoder:}~
        Taking the acoustic feature as input, the model generates the transcription by the first decoder, then combines the representation of the encoder and the first decoder to generate the translation~\cite{Sperber_CL2019}.
        However, two-pass decoding loses the inherent advantages of low latency due to sequential generation.
        To do this,~\citet{Inaguma_ASRU2021} utilize the non-autoregressive method for the first-stage decoding.
        
    \item \textbf{Dual decoder:}~
        Considering that generation processes of transcription and translation can help each other, 
        interactive decoding~\cite{Liu_AAAI2020,Le_COLING20202} is proposed to generate transcription and translation using two decoders synchronously.
        Also, an additional cross-attention module is used to capture the information from the two decoders to each other.
        To further improve translation performance, wait-k policy~\cite{ma2018stacl} is introduced, where transcribed tokens are first predicted, in order to provide more useful information for the decoding of the translation tokens.
\end{itemize}

\subsubsection{Decoupled Encoder}
\label{sec:model:mlt:decouple_enc}
Although the E2E model can use the decoupled decoder to partially alleviate the heavy modeling burden, multiple inferences usually lead to complex designs and high latency.
Instead of the decoupled decoder, a better choice is to simultaneously recognize and understand the semantics of the original speech input by a decoupled encoder.
As shown in Figure~\ref{fig:decouple_enc}, the decoupled encoder generally has two encoders --- the low-level speech encoder first encodes the acoustic information from the speech input, and the semantic encoder further learns the semantic representation needed for translation decoding. 

Each stage of encoding can be supervised by the information from the transcription, such as phonemes, text, etc.
Such decoupling mimics the cascaded system while transcription provides an alignment to the speech, which helps to ease the encoding burden.
For example, studies~\cite{Liu_corr2020,dong2021listen,Xu_ACL2021} decouple the encoder and add a Connectionist Temporal Classification (CTC) loss term after the speech encoder to predict the transcription.
\citet{Xu_ACL2021} analyze the different behavior of the self-attention in the encoders of ASR, MT and ST models with respect to the attention to local information, and demonstrate that for ST, the CTC module acts more efficiently over the low-level acoustic encoder. 
Given the length gap between speech and text, \citet{Liu_corr2020} propose a shrinking mechanism based on CTC loss and integrate with multitask learning.
In addition, \citet{dong2021listen} introduce an additional pre-trained BERT model
to supervise the output of the high-level acoustic encoder.

\subsubsection{Two-stream Encoder}

In decoupling methods, it is easy to further boost part of components by utilizing the additional ASR data.
However, having the potential to help the ST model improve the ability of semantic understanding, abundant knowledge from MT data is not exploited.
As shown in Figure~\ref{fig:2_enc}, the two-stream encoder is proposed to accept either speech or text or both as inputs during training.
Speech and text have both their separate encoders (\textit{speech encoder} and \textit{text encoder}, respectively) as the pre-net and a \textit{shared encoder} stacked on top. 
This structure is usually optimized by multitask training losses, such as NLL losses for both ST and MT~\cite{tang2021general,han2021learning,Ye_interspeech2021,tang2021improving,Ye_NAACL2022}.
Its advantage is that better semantic representations can be learned by sharing with the MT encoder to improve translation performance.
During inference, we input the speech, and the translated text is generated through the speech encoder, shared encoder, and decoder.

\begin{itemize}

\item \textbf{Speech encoder:}~
Speech encoder needs to be more capable to extract acoustic features of the speech input separately.
Pre-trained speech models such as Wav2vec2 can be used as the speech encoder for better ST performances~\cite{han2021learning,Ye_interspeech2021,tang2021improving,fang2022stemm,Ye_NAACL2022}.

\item \textbf{Text encoder:}~
Text encoder can be the text embedding layer or a few layers of the textual Transformer encoder.
\citet{tang2021general} propose to replace the original transcription with the phoneme of the speech as the text input, which helps to reduce the gap between two input modalities, thus improving performance.

\item \textbf{Interaction:}~
Under the overall two-steam encoder framework, there are also multiple variants of interaction between the speech encoder and the text encoder or their output representations.
\citet{Ye_NAACL2022} notice the sentence-level representation gap in the vector space between the speech and text, and propose to apply the contrastive learning method to draw close the two representations.
Starting from the length gap between the speech and text, \citet{han2021learning} propose the Chimera model, whose core idea is to align and map audio and text representations to the same length by a fix-size shared semantic memory module, and the decoder cross-attends to the memory module during autoregressive generation.
Considering both representations and the length difference, \citet{tang2021improving} add a cross-attentive regularization module after the shared encoder. The regularization module first generates two reconstructed sequences from text or speech encoders with the same length via self-attention or cross-attention, and then optimizes the L2 distance between the reconstructed sequences.
\citet{indurthi2021task} design a Task Characteristics Network (TCN) that produces a task embedding to modulate the parameters of the shared encoder-decoder. 

\end{itemize}

\subsection{Non-autoregressive Modeling}
\label{sec:model:nat}
E2E modeling reduces latency by almost half compared to cascaded counterpart, which helps applications in real-world scenarios with limited computational resources.
However, this advantage only holds in the context of autoregressive decoding, which generates each token depending on the previously predicted tokens.
The recently proposed non-autoregressive (NAR) decoding
predicts the whole sequence in parallel, eliminating the advantage of the E2E model in inference latency.

To combine E2E modeling and NAR generation, several studies explore NAR speech translation.
There are two design concepts.
Following the methods in ASR and MT tasks, \citet{Inaguma_ICASSP2021} combine multiple existing techniques, like conditional masked language model and re-scoring.
Another route explores a more efficient architecture that relies only on CTC for prediction \cite{Chuang_ACL2021}, which has the potential to achieve high speed-up.
However, the current non-autoregressive model is still inferior to the autoregressive counterpart with a large gap of about 2 $\sim$ 3 BLEU points.
We need more effort to develop a powerful NAR model with comparable performance.

\section{Tackling Data Scarcity}
\label{sec:training}

Because of the difficulty in accumulating data, the training data for ST is much less, compared to MT or ASR.
First, expanding the dataset and data augmentation are the most straightforward ideas~(Section \ref{sec:train:DA}).
Second, the majority of the existing research focuses on how to gain more knowledge or information from MT or ASR data or models to improve the performance of the ST model. We divide them into two parts: pre-training (Section \ref{sec:train:pretrain}) and knowledge distillation~(Section \ref{sec:train:KD}).
Finally, we present some progress on the current multilingual speech translation in Section \ref{sec:train:multilingual}.

\subsection{Data Augmentation}
\label{sec:train:DA}
\label{sec:data_aug}
Data augmentation is the most straightforward idea when training data is scarce.

\paragraph{Expanding ST data.}~
Intuitively, we can expand a large amount of target language translation by using a high-quality off-the-shelf MT system on a large amount of ASR data~\cite{Pino_ISCA2020,Gaido_IWSLT2020,wang2021large}.
This method is often referred to as pseudo-labeling or sequence-level knowledge distillation~(SeqKD).
\citet{inaguma2021source} propose bidirectional SeqKD that fully leverages knowledge in both source and target language directions (corresponding to forward SeqKD and backward SeqKD) for bilingual E2E-ST models. 
The source language speech can also be augmented, we can use the text-to-speech (TTS) model to extend the source-side text of MT into speech~\cite{jia2019leveraging}.

\paragraph{Speech augmentation.}~
Speech augmentation is also useful to improve performance as well as robustness, because speech is more complex and varied than text, and a single utterance can have different continuous signals depending on the speaker, recorder, environment, and so on.
SpecAugment~\cite{Park_ISCA2019} is commonly applied directly to the filter bank coefficients of speech inputs. The augmentation strategy contains warping features, masking blocks of frequency channels, and time steps. It has been proven to be effective on both ASR and ST tasks~\cite{bahar2019using}.
\citet{mccarthy2020skinaugment} propose SkinAugment, which uses auto-encoding speaker conversion to convert the original speaker's voice into another speaker's voice.
In addition, more diverse ST data can also be constructed by various segmentation methods~\cite{tsiamas2022segaugment} and recombination~\cite{lam2022sample} to enhance the utility of the original ST data.


\subsection{Pre-training}
\label{sec:train:pretrain}

On many tasks in the AI field, pre-training can greatly improve model performance in low-resource situations.
Pre-training is generally considered to have the following benefits:
(1) Compared to speech-to-translation corpora, the data used for pre-training is usually easy to obtain, such as a large amount of raw data of text sentences or speech.
The large-scale data used in pre-training (whether in-domain or out-of-domain) helps to improve the robustness of the model for the downstream tasks.
(2) Through basic and various pre-training tasks, such as reconstruction, mask-prediction, and contrastive learning, we can obtain more accurate representations with contextual information. These representations are generally helpful for various downstream tasks.

Pre-training is very effective in improving the performance of end-to-end speech translation, and throughout the state-of-the-art~(SOTA) E2E ST models, pre-training is always involved.
We outline two pre-training modes, namely \textit{separate pre-training} and \textit{joint pre-training}, based on the percentage of pre-trained modules in the E2E ST model.

\paragraph{Separate pre-training.}~
Separate pre-training refers to the pre-training of a portion of the model parameters or the pre-training of different sub-modules via different tasks.
The earlier work explores better pre-training methods to enhance the ability of the encoder in terms of semantic understanding. 
\citet{Wang_acl2020} pre-train the encoder using a curriculum learning method to improve syntactic and semantic modeling abilities.
\citet{Chen_Corr2020} propose a self-supervised method called Masked Acoustic Modeling (MAM), which randomly masks part of the speech spectrogram and then recovers it on top of the encoder.
\citet{Zheng_ICML2021} propose FAT based on MAM. It unifies speech and text representation through masked language modeling.
In addition, as discussed in Section \ref{sec:model:enc_dec}, the self-supervised model such as Wav2vec~\cite{Steffen_Interspeech2019} can act as a feature extractor~\cite{Nguyen_ISCA2020,Wu_ISCA2020} instead of random parameter initialization, providing effective acoustic features as input.

\paragraph{Joint pre-training.}~
Joint pre-training means that the model (all modules including both encoder and decoder) participates in pre-training as a whole.
Joint pre-training usually enjoys a multitask learning framework, which is introduced in Section~\ref{sec:model:mlt}).
In the multitask pre-training framework, unified models are built to jointly pre-train ASR, MT, masked language modeling, or even speech (re-)synthesis tasks, using speech-text supervised data as well as large amounts of unlabeled text and speech.
After pre-training, the models only need to be fine-tuned with the speech-translated parallel corpus to achieve a decent result.
For instance,
\citet{ao2022speecht5} propose SpeechT5, which pre-trains various speech/text-to-speech/text tasks, including ASR, ST, text-to-speech, speech conversion, and speech enhancement.
\citet{tang2022unified} incorporate both self-supervised and supervised pre-training tasks, including (self-)supervised text-to-text, speech SSL learning, speech-to-phoneme, ASR, and ST. 
In addition, pre-training can also be combined with multilingualism. 
\citet{bapna2022mslam} propose mSLAM, a large multilingual speech-text Conformer model based on the two-stream encoder, which surprisingly shows some zero-shot learning capability.
\citet{cheng2022mu} further combine mSLAM with multitask learning, propose Mu2SLAM and obtain the SOTA results on CoVoST-2~\cite{wang2020covost} multilingual ST benchmark.

\subsection{Knowledge Distillation}
\label{sec:train:KD}
\label{sec:kd}
Knowledge distillation (KD)
is typically used for model compression, using the output of a larger teacher model that typically performs better to guide the learning of a student model, with the expectation that the student model will achieve the same performance as the teacher model.
With limited data, how can we get the ST performance close to that of the MT teacher? The idea of knowledge distillation is then widely used in speech translation.
A straightforward approach is to use the ST model and the MT model to predict translation tokens separately, with the prediction probabilities of the MT model serving as the teacher to guide the ST output~\cite{Liu_ISCA2019,tang2021improving}.
With a two-stream encoder framework, \citet{fang2022stemm} propose to distill the speech-to-text translation module with the translation output from the speech-text manifold mix-up sequence. Experiments show that the mix-up sequence can bridge the representation gap between speech and text, thus making the learned semantic representation of text more readily transferable to speech.


\subsection{Multilingual Training}
\label{sec:train:multilingual}
Multilingual speech translation includes one-to-many, many-to-one, and many-to-many scenarios.
Like MT, adding language indicators, such as \texttt{<2de>}, \texttt{<2fr>}, to the decoder is the most straightforward and effective way to evolve from bilingual to multilingual ST~\cite{inaguma2019multilingual}.
\citet{wang2020covost} also show that with limited data for each translation direction, training a many-to-many multilingual ST model is better than training bilingual ST models individually, because the multilingual model can capture more pronunciation similarity between languages.
Current research on multilingual ST mainly focuses on pre-training, such as how to build a unified multilingual speech-text pre-training model~\cite{bapna2022mslam} and how to design various and effective pre-training tasks~\cite{cheng2022mu}. These models can be helpful for translation as well as multilingual ASR.
There is also some work focusing on efficient fine-tuning.
\citet{li2021multilingual} concatenate multilingual pre-trained XLSR speech encoder
~\cite{conneau2020unsupervised} 
with mBART
~\cite{liu2020multilingual} 
decoder and experimentally find that fine-tuning the parameters of layer-norm and attention layers is better than fine-tuning all parameters.
\citet{le2021lightweight}, on the other hand, freeze the pre-trained ASR encoder and the mBART decoder, and complete one-to-many ST by only tuning the language-specific adapter modules on top of a multilingual system, with only tens of millions of parameters.


\section{Tackling Application Issues}
\label{sec:application}

Current research is usually conducted under presupposed settings with manual segmentation, noise-free environment, etc. And the demands of practical application are rarely discussed. 

\paragraph{Real-time.}~
Simultaneous decoding aims for the quality-latency trade-off for real-time translation on the depend of a decision policy, which determines whether to wait for more audio stream or decode one or more tokens. 
SimulSpeech~\cite{ren2020simulspeech} propose a speech segmenter, based on the CTC criterion, to split the streaming speech in real time.~\citet{chang2022exploring} adapt Continuous Integrate-and-Fire module
to play a role as the adaptive policy, which makes WRITE decisions at each firing step.
\citet{liu2021cross} extend RNN-Transducer
into Cross Attention Augmented Transducer, which can jointly optimize the decoding policy and translation quality by considering all possible READ and WRITE action paths. 

\paragraph{Segmentation.}~
The direct ST model cannot handle long audios alone, such as a complete speech or a movie, which have to be segmented into shorter utterances using automatic segmentation methods (VAD-based, fixed-length, and hybrid).
However, there exists a gap between manual segmentation during training and automatic segmentation at runtime.   
\citet{tsiamas2022shas} propose Supervised Hybrid Audio Segmentation (SHAS) method, which uses Wav2vec2 and trains a classifier to predict the split locations supervised by the manual segmentation information. 
\citet{gaido2021beyond} propose an enhanced hybrid method considering both the audio's pauses and the target segments' length.  

\paragraph{Named entity.}~
How to handle the translation of named entity (NE) is a critical demand for ST systems in real-world scenarios. 
~\citet{gaido2022we} discover that the nationality of the referred person is the key factor for the failures in person name translation, and propose multilingual models to increase the robustness of varied pronunciations.
\citet{gaido2022joint} design two methods to jointly perform ST and recognize NE, of which the \textit{inline} method generates NE tags and tokens successively, while the \textit{parallel} method predicts NE tags and tokens in parallel with two linear layers. 

\paragraph{Code-switching.}~
Code-switching (CS) speech commonly exists in casual situations, and as blending different languages, translating CS speech is a challenge.
\citet{weller2022end} create a CS corpus and explore
both cascaded and end-to-end architecture to perform CS speech translation.
\citet{huber2022code} propose a unified Language Agnostic E2E ST model~(LAST) by training both ASR and ST tasks, as well as enlarging CS data through concatenation.

\paragraph{Gender bias.}~
Addressing gender bias in translation is a relatively new area of NLP and speech research.
For ST task, audio input contains more clues about gender identity.
\citet{bentivogli2020gender} release MuST-SHE benchmark allowing for the fine-grained analysis of gender bias in ST.
They also find that the end-to-end approach can directly use audio information and have more potential to better address gender issues.
\citet{savoldi2022under} later extend MuST-SHE with two additional linguistic information, part-of-speech and agreement chains.
\citet{gaido2021split} investigate how segmentation methods influence the translation of gender, and propose a combined segmentation method with both subword splitting and character-based splitting.




\section{Future}
\label{sec:future}


In this paper, we thoroughly present recent advances in direct speech-to-text translation. 
Specifically, we review and summarize existing approaches in this field for the first time with an original taxonomy. 
Despite the recent attractive progress of direct ST technology, there remain many unresolved problems to be explored. 
Finally, we discuss some promising topics for the future.

\paragraph{LLM.}~
Today, large language models (LLMs), such as ChatGPT, Bloom~\cite{scao2022bloom}, have shown powerful dexterity in a variety of NLP applications, such as text generation and even in the zero-shot scenario.
First of all, we believe that it is worthwhile to further explore how to integrate the powerful generative capabilities of LLMs into ST tasks and to incorporate speech data into the training of LLMs.
As an initial step, for instance, we may optimize speech representation to be comparable to the text representation as a prompt function to interact with LLMs. We conjecture that speech discrete representations as pseudo-language~\cite{hsu2021hubert,wu2022wav2seq} may be interesting prompts.
Furthermore, pre-training large-scale acoustics-aware LLMs is also a promising direction that will greatly promote the entire NLP and speech community. We anticipate that after scaling up further, the models will have the capability for few-shot ST, zero-shot ST, and transfer learning.

\paragraph{Multimodality.}~
Numerous human-computer interaction (HCI) application scenarios have emerged with the recent worldwide surge in AI-generated content (text, images, voice, and video, etc.), which drives the ST field to explore more sophisticated directions, like speech-to-speech translation and video translation.
With the explosive growth of multimodal resources, how to perform in-context learning (ICL) on multimodal data is also a promising research topic.
Recently, multimodal pre-training has already been proved to be effective in many fields, such as Data2Vec~\cite{baevski2022data2vec}.
However, the interactions and interrelated information between multiple modalities (e.g., the speech of characters in videos and their corresponding image frames and prosodic environments) remain underutilized.
We believe that a more unified and robust pre-training paradigm, aimed at learning universal cross-lingual cross-modal representations, is important for ST and the more demanding HCI scenarios mentioned above.




\section*{Acknowledgements}

This work was supported in part by the National Science Foundation of China (No. 62276056), the National Key R\&D Program of China, the China HTRD Center Project (No. 2020AAA0107904), the Natural Science Foundation of Liaoning Province of China (2022-KF-16-01), the Yunnan Provincial Major Science and Technology Special Plan Projects (No. 202103AA080015), the Fundamental Research Funds for the Central Universities (Nos. N2216016, N2216001, and N2216002), and the Program of Introducing Talents of Discipline to Universities, Plan 111 (No. B16009).

\section*{Contribution Statement}

\noindent \textbf{Chen Xu}, \textbf{Rong Ye}, and \textbf{Qianqian Dong} have the equal contribution to the conception and design of the survey.
They performed the literature search, collected relevant papers, and drafted the main sections of the manuscript. 
They also contributed to the synthesis and analysis of the surveyed work and were responsible for creating the figures.

\textbf{Chengqi Zhao} provided substantial contributions to the refinement of the survey's scope and focus. 
He helped in the analysis of the surveyed work and provided insights on potential future research directions. 
In addition, he was responsible for critically reviewing the manuscript at different stages and provided suggestions for improving the clarity and coherence of the paper.

\textbf{Tom Ko}, \textbf{Mingxuan Wang}, \textbf{Tong Xiao}, and \textbf{Jingbo Zhu} provided guidance on the overall direction and structure of the survey paper. 
They contributed to the refinement of the paper's scope and focus, and provided critical feedback on the manuscript at various stages of development. They also helped in the identification of relevant literature and offered insights on the state-of-the-art and future trends in end-to-end speech translation.

\bibliographystyle{named}
\bibliography{rebib,new}

\begin{thebibliography}{}

\bibitem[\protect\citeauthoryear{Anastasopoulos and
  Chiang}{2018}]{Anastasopoulos_NAACL2018}
Antonios Anastasopoulos and David Chiang.
\newblock Tied multitask learning for neural speech translation.
\newblock In {\em NAACL}, 2018.

\bibitem[\protect\citeauthoryear{Ansari \bgroup \em et al.\egroup
  }{2020}]{Anastasopoulos_IWSLT2022}
Ebrahim Ansari, Amittai Axelrod, Nguyen Bach, Ond{\v{r}}ej Bojar, Roldano
  Cattoni, Fahim Dalvi, Nadir Durrani, Marcello Federico, Christian Federmann,
  Jiatao Gu, Fei Huang, Kevin Knight, Xutai Ma, Ajay Nagesh, Matteo Negri, Jan
  Niehues, Juan Pino, Elizabeth Salesky, Xing Shi, Sebastian St{\"u}ker, Marco
  Turchi, Alexander Waibel, and Changhan Wang.
\newblock Findings of the iwslt 2020 evaluation campaign.
\newblock In {\em IWSLT}, 2020.

\bibitem[\protect\citeauthoryear{Ao \bgroup \em et al.\egroup
  }{2022}]{ao2022speecht5}
Junyi Ao, Rui Wang, Long Zhou, Chengyi Wang, Shuo Ren, Yu~Wu, Shujie Liu, Tom
  Ko, Qing Li, Yu~Zhang, Zhihua Wei, Yao Qian, Jinyu Li, and Furu Wei.
\newblock {S}peech{T}5: Unified-modal encoder-decoder pre-training for spoken
  language processing.
\newblock In {\em ACL}, 2022.

\bibitem[\protect\citeauthoryear{Baevski \bgroup \em et al.\egroup
  }{2020}]{baevski2020wav2vec}
Alexei Baevski, Yuhao Zhou, Abdelrahman Mohamed, and Michael Auli.
\newblock wav2vec 2.0: {A} framework for self-supervised learning of speech
  representations.
\newblock In {\em NeurIPS}, 2020.

\bibitem[\protect\citeauthoryear{Baevski \bgroup \em et al.\egroup
  }{2022}]{baevski2022data2vec}
Alexei Baevski, Wei-Ning Hsu, Qiantong Xu, Arun Babu, Jiatao Gu, and Michael
  Auli.
\newblock Data2vec: A general framework for self-supervised learning in speech,
  vision and language.
\newblock In {\em International Conference on Machine Learning}, pages
  1298--1312. PMLR, 2022.

\bibitem[\protect\citeauthoryear{Bahar \bgroup \em et al.\egroup
  }{2019}]{bahar2019using}
Parnia Bahar, Albert Zeyer, Ralf Schl{\"u}ter, and Hermann Ney.
\newblock On using {S}pec{A}ugment for end-to-end speech translation.
\newblock In {\em IWSLT}, 2019.

\bibitem[\protect\citeauthoryear{Bapna \bgroup \em et al.\egroup
  }{2022}]{bapna2022mslam}
Ankur Bapna, Colin Cherry, Yu~Zhang, Ye~Jia, Melvin Johnson, Yong Cheng, Simran
  Khanuja, Jason Riesa, and Alexis Conneau.
\newblock mslam: Massively multilingual joint pre-training for speech and text.
\newblock {\em CoRR}, 2022.

\bibitem[\protect\citeauthoryear{Bentivogli \bgroup \em et al.\egroup
  }{2020}]{bentivogli2020gender}
Luisa Bentivogli, Beatrice Savoldi, Matteo Negri, Mattia~A. Di~Gangi, Roldano
  Cattoni, and Marco Turchi.
\newblock Gender in danger? evaluating speech translation technology on the
  {M}u{ST}-{SHE} corpus.
\newblock In {\em ACL}, 2020.

\bibitem[\protect\citeauthoryear{Bentivogli \bgroup \em et al.\egroup
  }{2021}]{Bentivogli_ACL2021}
Luisa Bentivogli, Mauro Cettolo, Marco Gaido, Alina Karakanta, Alberto
  Martinelli, Matteo Negri, and Marco Turchi.
\newblock Cascade versus direct speech translation: Do the differences still
  make a difference?
\newblock In {\em ACL}, 2021.

\bibitem[\protect\citeauthoryear{B{\'e}rard \bgroup \em et al.\egroup
  }{2016}]{Berard_arxiv2016}
Alexandre B{\'e}rard, Olivier Pietquin, Laurent Besacier, and Christophe
  Servan.
\newblock Listen and translate: A proof of concept for end-to-end
  speech-to-text translation.
\newblock In {\em NIPS Workshop on end-to-end learning for speech and audio
  processing}, 2016.

\bibitem[\protect\citeauthoryear{Chang and Lee}{2022}]{chang2022exploring}
Chih-Chiang Chang and Hung-yi Lee.
\newblock Exploring continuous integrate-and-fire for adaptive simultaneous
  speech translation.
\newblock {\em CoRR}, 2022.

\bibitem[\protect\citeauthoryear{Chen \bgroup \em et al.\egroup
  }{2020}]{Chen_Corr2020}
Junkun Chen, Mingbo Ma, Renjie Zheng, and Liang Huang.
\newblock {MAM:} masked acoustic modeling for end-to-end speech-to-text
  translation.
\newblock {\em CoRR}, 2020.

\bibitem[\protect\citeauthoryear{Cheng \bgroup \em et al.\egroup
  }{2022}]{cheng2022mu}
Yong Cheng, Yu~Zhang, Melvin Johnson, Wolfgang Macherey, and Ankur Bapna.
\newblock Mu2slam: Multitask, multilingual speech and language models.
\newblock {\em CoRR}, 2022.

\bibitem[\protect\citeauthoryear{Chuang \bgroup \em et al.\egroup
  }{2021}]{Chuang_ACL2021}
Shun-Po Chuang, Yung-Sung Chuang, Chih-Chiang Chang, and Hung-yi Lee.
\newblock Investigating the reordering capability in {CTC}-based
  non-autoregressive end-to-end speech translation.
\newblock In {\em Findings of ACL}, 2021.

\bibitem[\protect\citeauthoryear{Conneau \bgroup \em et al.\egroup
  }{2021}]{conneau2020unsupervised}
Alexis Conneau, Alexei Baevski, Ronan Collobert, Abdelrahman Mohamed, and
  Michael Auli.
\newblock Unsupervised cross-lingual representation learning for speech
  recognition.
\newblock In {\em InterSpeech}, 2021.

\bibitem[\protect\citeauthoryear{Di~Gangi \bgroup \em et al.\egroup
  }{2019}]{Gangi_NAACL2019}
Mattia~A. Di~Gangi, Roldano Cattoni, Luisa Bentivogli, Matteo Negri, and Marco
  Turchi.
\newblock {M}u{ST}-{C}: a {M}ultilingual {S}peech {T}ranslation {C}orpus.
\newblock In {\em NAACL}, 2019.

\bibitem[\protect\citeauthoryear{Dong \bgroup \em et al.\egroup
  }{2021}]{dong2021listen}
Qianqian Dong, Rong Ye, Mingxuan Wang, Hao Zhou, Shuang Xu, Bo~Xu, and Lei Li.
\newblock Listen, understand and translate: Triple supervision decouples
  end-to-end speech-to-text translation.
\newblock In {\em AAAI}, 2021.

\bibitem[\protect\citeauthoryear{Fang \bgroup \em et al.\egroup
  }{2022}]{fang2022stemm}
Qingkai Fang, Rong Ye, Lei Li, Yang Feng, and Mingxuan Wang.
\newblock {STEMM}: Self-learning with speech-text manifold mixup for speech
  translation.
\newblock In {\em ACL}, 2022.

\bibitem[\protect\citeauthoryear{Gaido \bgroup \em et al.\egroup
  }{2020}]{Gaido_IWSLT2020}
Marco Gaido, Mattia~A. Di~Gangi, Matteo Negri, and Marco Turchi.
\newblock End-to-end speech-translation with knowledge distillation:
  {FBK}@{IWSLT}2020.
\newblock In {\em IWSLT}, 2020.

\bibitem[\protect\citeauthoryear{Gaido \bgroup \em et al.\egroup
  }{2021a}]{gaido2021beyond}
Marco Gaido, Matteo Negri, Mauro Cettolo, and Marco Turchi.
\newblock Beyond voice activity detection: Hybrid audio segmentation for direct
  speech translation.
\newblock In {\em ICNLSP}, 2021.

\bibitem[\protect\citeauthoryear{Gaido \bgroup \em et al.\egroup
  }{2021b}]{gaido2021split}
Marco Gaido, Beatrice Savoldi, Luisa Bentivogli, Matteo Negri, and Marco
  Turchi.
\newblock How to split: the effect of word segmentation on gender bias in
  speech translation.
\newblock In {\em Findings of ACL}, 2021.

\bibitem[\protect\citeauthoryear{Gaido \bgroup \em et al.\egroup
  }{2022a}]{gaido2022we}
Marco Gaido, Matteo Negri, and Marco Turchi.
\newblock Who are we talking about? handling person names in speech
  translation.
\newblock In {\em IWSLT}, 2022.

\bibitem[\protect\citeauthoryear{Gaido \bgroup \em et al.\egroup
  }{2022b}]{gaido2022joint}
Marco Gaido, Sara Papi, Matteo Negri, and Marco Turchi.
\newblock Joint speech translation and named entity recognition.
\newblock {\em CoRR}, 2022.

\bibitem[\protect\citeauthoryear{Gangi \bgroup \em et al.\egroup
  }{2019}]{di2019adapting}
Mattia Antonino~Di Gangi, Matteo Negri, and Marco Turchi.
\newblock Adapting transformer to end-to-end spoken language translation.
\newblock In {\em InterSpeech}, 2019.

\bibitem[\protect\citeauthoryear{Gulati \bgroup \em et al.\egroup
  }{2020}]{Gulati_ISCA2020}
Anmol Gulati, James Qin, Chung{-}Cheng Chiu, Niki Parmar, Yu~Zhang, Jiahui Yu,
  Wei Han, Shibo Wang, Zhengdong Zhang, Yonghui Wu, and Ruoming Pang.
\newblock Conformer: Convolution-augmented transformer for speech recognition.
\newblock In {\em InterSpeech}, 2020.

\bibitem[\protect\citeauthoryear{Han \bgroup \em et al.\egroup
  }{2021}]{han2021learning}
Chi Han, Mingxuan Wang, Heng Ji, and Lei Li.
\newblock Learning shared semantic space for speech-to-text translation.
\newblock In {\em Findings of ACL}, 2021.

\bibitem[\protect\citeauthoryear{Hsu \bgroup \em et al.\egroup
  }{2021}]{hsu2021hubert}
Wei-Ning Hsu, Benjamin Bolte, Yao-Hung~Hubert Tsai, Kushal Lakhotia, Ruslan
  Salakhutdinov, and Abdelrahman Mohamed.
\newblock Hubert: Self-supervised speech representation learning by masked
  prediction of hidden units.
\newblock {\em TASLP}, 2021.

\bibitem[\protect\citeauthoryear{Huber \bgroup \em et al.\egroup
  }{2022}]{huber2022code}
Christian Huber, Enes~Yavuz Ugan, and Alexander Waibel.
\newblock Code-switching without switching: Language agnostic end-to-end speech
  translation.
\newblock {\em CoRR}, 2022.

\bibitem[\protect\citeauthoryear{Inaguma \bgroup \em et al.\egroup
  }{2019}]{inaguma2019multilingual}
Hirofumi Inaguma, Kevin Duh, Tatsuya Kawahara, and Shinji Watanabe.
\newblock Multilingual end-to-end speech translation.
\newblock In {\em ASRU}, 2019.

\bibitem[\protect\citeauthoryear{Inaguma \bgroup \em et al.\egroup
  }{2021a}]{Inaguma_ASRU2021}
Hirofumi Inaguma, Siddharth Dalmia, Brian Yan, and Shinji Watanabe.
\newblock Fast-md: Fast multi-decoder end-to-end speech translation with
  non-autoregressive hidden intermediates.
\newblock In {\em ASRU}, 2021.

\bibitem[\protect\citeauthoryear{Inaguma \bgroup \em et al.\egroup
  }{2021b}]{Inaguma_ICASSP2021}
Hirofumi Inaguma, Yosuke Higuchi, Kevin Duh, Tatsuya Kawahara, and Shinji
  Watanabe.
\newblock {ORTHROS:} non-autoregressive end-to-end speech translation with
  dual-decoder.
\newblock In {\em ICASSP}, 2021.

\bibitem[\protect\citeauthoryear{Inaguma \bgroup \em et al.\egroup
  }{2021c}]{inaguma2021source}
Hirofumi Inaguma, Tatsuya Kawahara, and Shinji Watanabe.
\newblock Source and target bidirectional knowledge distillation for end-to-end
  speech translation.
\newblock In {\em NAACL}, 2021.

\bibitem[\protect\citeauthoryear{Indurthi \bgroup \em et al.\egroup
  }{2021}]{indurthi2021task}
Sathish Indurthi, Mohd~Abbas Zaidi, Nikhil~Kumar Lakumarapu, Beomseok Lee,
  Hyojung Han, Seokchan Ahn, Sangha Kim, Chanwoo Kim, and Inchul Hwang.
\newblock Task aware multi-task learning for speech to text tasks.
\newblock In {\em ICASSP}, 2021.

\bibitem[\protect\citeauthoryear{Jia \bgroup \em et al.\egroup
  }{2019}]{jia2019leveraging}
Ye~Jia, Melvin Johnson, Wolfgang Macherey, Ron~J. Weiss, Yuan Cao,
  Chung{-}Cheng Chiu, Naveen Ari, Stella Laurenzo, and Yonghui Wu.
\newblock Leveraging weakly supervised data to improve end-to-end
  speech-to-text translation.
\newblock In {\em ICASSP}, 2019.

\bibitem[\protect\citeauthoryear{Lam \bgroup \em et al.\egroup
  }{2022}]{lam2022sample}
Tsz~Kin Lam, Shigehiko Schamoni, and Stefan Riezler.
\newblock Sample, translate, recombine: Leveraging audio alignments for data
  augmentation in end-to-end speech translation.
\newblock In {\em ACL}, 2022.

\bibitem[\protect\citeauthoryear{Le \bgroup \em et al.\egroup
  }{2020}]{Le_COLING20202}
Hang Le, Juan Pino, Changhan Wang, Jiatao Gu, Didier Schwab, and Laurent
  Besacier.
\newblock Dual-decoder transformer for joint automatic speech recognition and
  multilingual speech translation.
\newblock In {\em COLING}, 2020.

\bibitem[\protect\citeauthoryear{Le \bgroup \em et al.\egroup
  }{2021}]{le2021lightweight}
Hang Le, Juan Pino, Changhan Wang, Jiatao Gu, Didier Schwab, and Laurent
  Besacier.
\newblock Lightweight adapter tuning for multilingual speech translation.
\newblock In {\em ACL}, 2021.

\bibitem[\protect\citeauthoryear{Li \bgroup \em et al.\egroup
  }{2021}]{li2021multilingual}
Xian Li, Changhan Wang, Yun Tang, Chau Tran, Yuqing Tang, Juan Pino, Alexei
  Baevski, Alexis Conneau, and Michael Auli.
\newblock Multilingual speech translation from efficient finetuning of
  pretrained models.
\newblock In {\em ACL}, 2021.

\bibitem[\protect\citeauthoryear{Liu \bgroup \em et al.\egroup
  }{2019}]{Liu_ISCA2019}
Yuchen Liu, Hao Xiong, Jiajun Zhang, Zhongjun He, Hua Wu, Haifeng Wang, and
  Chengqing Zong.
\newblock End-to-end speech translation with knowledge distillation.
\newblock In {\em InterSpeech}, 2019.

\bibitem[\protect\citeauthoryear{Liu \bgroup \em et al.\egroup
  }{2020a}]{liu2020multilingual}
Yinhan Liu, Jiatao Gu, Naman Goyal, Xian Li, Sergey Edunov, Marjan
  Ghazvininejad, Mike Lewis, and Luke Zettlemoyer.
\newblock Multilingual denoising pre-training for neural machine translation.
\newblock {\em TACL}, 2020.

\bibitem[\protect\citeauthoryear{Liu \bgroup \em et al.\egroup
  }{2020b}]{Liu_AAAI2020}
Yuchen Liu, Jiajun Zhang, Hao Xiong, Long Zhou, Zhongjun He, Hua Wu, Haifeng
  Wang, and Chengqing Zong.
\newblock Synchronous speech recognition and speech-to-text translation with
  interactive decoding.
\newblock In {\em AAAI}, 2020.

\bibitem[\protect\citeauthoryear{Liu \bgroup \em et al.\egroup
  }{2020c}]{Liu_corr2020}
Yuchen Liu, Junnan Zhu, Jiajun Zhang, and Chengqing Zong.
\newblock Bridging the modality gap for speech-to-text translation.
\newblock {\em CoRR}, 2020.

\bibitem[\protect\citeauthoryear{Liu \bgroup \em et al.\egroup
  }{2021}]{liu2021cross}
Dan Liu, Mengge Du, Xiaoxi Li, Ya~Li, and Enhong Chen.
\newblock Cross attention augmented transducer networks for simultaneous
  translation.
\newblock In {\em EMNLP}, 2021.

\bibitem[\protect\citeauthoryear{Ma \bgroup \em et al.\egroup
  }{2019}]{ma2018stacl}
Mingbo Ma, Liang Huang, Hao Xiong, Renjie Zheng, Kaibo Liu, Baigong Zheng,
  Chuanqiang Zhang, Zhongjun He, Hairong Liu, Xing Li, Hua Wu, and Haifeng
  Wang.
\newblock {STACL}: Simultaneous translation with implicit anticipation and
  controllable latency using prefix-to-prefix framework.
\newblock In {\em ACL}, 2019.

\bibitem[\protect\citeauthoryear{McCarthy \bgroup \em et al.\egroup
  }{2020}]{mccarthy2020skinaugment}
Arya~D. McCarthy, Liezl Puzon, and Juan Pino.
\newblock Skinaugment: Auto-encoding speaker conversions for automatic speech
  translation.
\newblock In {\em ICASSP}, 2020.

\bibitem[\protect\citeauthoryear{Nguyen \bgroup \em et al.\egroup
  }{2020}]{Nguyen_ISCA2020}
Ha~Nguyen, Fethi Bougares, Natalia~A. Tomashenko, Yannick Est{\`{e}}ve, and
  Laurent Besacier.
\newblock Investigating self-supervised pre-training for end-to-end speech
  translation.
\newblock In {\em InterSpeech}, 2020.

\bibitem[\protect\citeauthoryear{Panayotov \bgroup \em et al.\egroup
  }{2015}]{Panayotov_ICASSP2015}
Vassil Panayotov, Guoguo Chen, Daniel Povey, and Sanjeev Khudanpur.
\newblock Librispeech: An {ASR} corpus based on public domain audio books.
\newblock In {\em ICASSP}, 2015.

\bibitem[\protect\citeauthoryear{Park \bgroup \em et al.\egroup
  }{2019}]{Park_ISCA2019}
Daniel~S. Park, William Chan, Yu~Zhang, Chung{-}Cheng Chiu, Barret Zoph,
  Ekin~D. Cubuk, and Quoc~V. Le.
\newblock Specaugment: {A} simple data augmentation method for automatic speech
  recognition.
\newblock In {\em InterSpeech}, 2019.

\bibitem[\protect\citeauthoryear{Pino \bgroup \em et al.\egroup
  }{2020}]{Pino_ISCA2020}
Juan~Miguel Pino, Qiantong Xu, Xutai Ma, Mohammad~Javad Dousti, and Yun Tang.
\newblock Self-training for end-to-end speech translation.
\newblock In {\em InterSpeech}, 2020.

\bibitem[\protect\citeauthoryear{Ren \bgroup \em et al.\egroup
  }{2020}]{ren2020simulspeech}
Yi~Ren, Jinglin Liu, Xu~Tan, Chen Zhang, Tao Qin, Zhou Zhao, and Tie-Yan Liu.
\newblock {S}imul{S}peech: End-to-end simultaneous speech to text translation.
\newblock In {\em ACL}, 2020.

\bibitem[\protect\citeauthoryear{Savoldi \bgroup \em et al.\egroup
  }{2022}]{savoldi2022under}
Beatrice Savoldi, Marco Gaido, Luisa Bentivogli, Matteo Negri, and Marco
  Turchi.
\newblock Under the morphosyntactic lens: A multifaceted evaluation of gender
  bias in speech translation.
\newblock In {\em ACL}, 2022.

\bibitem[\protect\citeauthoryear{Scao \bgroup \em et al.\egroup
  }{2022}]{scao2022bloom}
Teven~Le Scao, Angela Fan, Christopher Akiki, Ellie Pavlick, Suzana Ili{\'c},
  Daniel Hesslow, Roman Castagn{\'e}, Alexandra~Sasha Luccioni, Fran{\c{c}}ois
  Yvon, Matthias Gall{\'e}, et~al.
\newblock Bloom: A 176b-parameter open-access multilingual language model.
\newblock {\em CoRR}, 2022.

\bibitem[\protect\citeauthoryear{Schneider \bgroup \em et al.\egroup
  }{2019}]{Steffen_Interspeech2019}
Steffen Schneider, Alexei Baevski, Ronan Collobert, and Michael Auli.
\newblock wav2vec: Unsupervised pre-training for speech recognition.
\newblock In {\em InterSpeech}, 2019.

\bibitem[\protect\citeauthoryear{Sperber and Paulik}{2020}]{sperber2020speech}
Matthias Sperber and Matthias Paulik.
\newblock Speech translation and the end-to-end promise: Taking stock of where
  we are.
\newblock In {\em ACL}, 2020.

\bibitem[\protect\citeauthoryear{Sperber \bgroup \em et al.\egroup
  }{2019}]{Sperber_CL2019}
Matthias Sperber, Graham Neubig, Jan Niehues, and Alex Waibel.
\newblock Attention-passing models for robust and data-efficient end-to-end
  speech translation.
\newblock {\em TACL}, 2019.

\bibitem[\protect\citeauthoryear{Stentiford and
  Steer}{1988}]{stentiford1988machine}
Fred~WM Stentiford and Martin~G Steer.
\newblock Machine translation of speech.
\newblock {\em British Telecom technology journal}, (2), 1988.

\bibitem[\protect\citeauthoryear{Tang \bgroup \em et al.\egroup
  }{2021a}]{tang2021improving}
Yun Tang, Juan Pino, Xian Li, Changhan Wang, and Dmitriy Genzel.
\newblock Improving speech translation by understanding and learning from the
  auxiliary text translation task.
\newblock In {\em ACL}, 2021.

\bibitem[\protect\citeauthoryear{Tang \bgroup \em et al.\egroup
  }{2021b}]{tang2021general}
Yun Tang, Juan Pino, Changhan Wang, Xutai Ma, and Dmitriy Genzel.
\newblock A general multi-task learning framework to leverage text data for
  speech to text tasks.
\newblock In {\em ICASSP}, 2021.

\bibitem[\protect\citeauthoryear{Tang \bgroup \em et al.\egroup
  }{2022}]{tang2022unified}
Yun Tang, Hongyu Gong, Ning Dong, Changhan Wang, Wei-Ning Hsu, Jiatao Gu,
  Alexei Baevski, Xian Li, Abdelrahman Mohamed, Michael Auli, and Juan Pino.
\newblock Unified speech-text pre-training for speech translation and
  recognition.
\newblock In {\em ACL}, 2022.

\bibitem[\protect\citeauthoryear{Tsiamas \bgroup \em et al.\egroup
  }{2022a}]{tsiamas2022segaugment}
Ioannis Tsiamas, Jos{\'e}~AR Fonollosa, and Marta~R Costa-juss{\`a}.
\newblock Segaugment: Maximizing the utility of speech translation data with
  segmentation-based augmentations.
\newblock {\em CoRR}, 2022.

\bibitem[\protect\citeauthoryear{Tsiamas \bgroup \em et al.\egroup
  }{2022b}]{tsiamas2022shas}
Ioannis Tsiamas, Gerard~I G{\'a}llego, Jos{\'e}~AR Fonollosa, and Marta~R
  Costa-juss{\`a}.
\newblock Shas: Approaching optimal segmentation for end-to-end speech
  translation.
\newblock In {\em InterSpeech}, 2022.

\bibitem[\protect\citeauthoryear{Vaswani \bgroup \em et al.\egroup
  }{2017}]{Vaswani_nips2017}
Ashish Vaswani, Noam Shazeer, Niki Parmar, Jakob Uszkoreit, Llion Jones,
  Aidan~N. Gomez, Lukasz Kaiser, and Illia Polosukhin.
\newblock Attention is all you need.
\newblock In {\em Neurips}, 2017.

\bibitem[\protect\citeauthoryear{Wang \bgroup \em et al.\egroup
  }{2020a}]{wang2020covost}
Changhan Wang, Juan Pino, Anne Wu, and Jiatao Gu.
\newblock {C}o{V}o{ST}: A diverse multilingual speech-to-text translation
  corpus.
\newblock In {\em IREC}, 2020.

\bibitem[\protect\citeauthoryear{Wang \bgroup \em et al.\egroup
  }{2020b}]{Wang_acl2020}
Chengyi Wang, Yu~Wu, Shujie Liu, Ming Zhou, and Zhenglu Yang.
\newblock Curriculum pre-training for end-to-end speech translation.
\newblock In {\em ACL}, 2020.

\bibitem[\protect\citeauthoryear{Wang \bgroup \em et al.\egroup
  }{2021}]{wang2021large}
Changhan Wang, Anne Wu, Juan Pino, Alexei Baevski, Michael Auli, and Alexis
  Conneau.
\newblock Large-scale self- and semi-supervised learning for speech
  translation.
\newblock In {\em InterSpeech}, 2021.

\bibitem[\protect\citeauthoryear{Weiss \bgroup \em et al.\egroup
  }{2017}]{Weiss_ISCA2017}
Ron~J. Weiss, Jan Chorowski, Navdeep Jaitly, Yonghui Wu, and Zhifeng Chen.
\newblock Sequence-to-sequence models can directly translate foreign speech.
\newblock In {\em InterSpeech}, 2017.

\bibitem[\protect\citeauthoryear{Weller \bgroup \em et al.\egroup
  }{2022}]{weller2022end}
Orion Weller, Matthias Sperber, Telmo Pires, Hendra Setiawan, Christian Gollan,
  Dominic Telaar, and Matthias Paulik.
\newblock End-to-end speech translation for code switched speech.
\newblock In {\em Findings of ACL}, 2022.

\bibitem[\protect\citeauthoryear{Wu \bgroup \em et al.\egroup
  }{2020}]{Wu_ISCA2020}
Anne Wu, Changhan Wang, Juan~Miguel Pino, and Jiatao Gu.
\newblock Self-supervised representations improve end-to-end speech
  translation.
\newblock In {\em InterSpeech}, 2020.

\bibitem[\protect\citeauthoryear{Wu \bgroup \em et al.\egroup
  }{2022}]{wu2022wav2seq}
Felix Wu, Kwangyoun Kim, Shinji Watanabe, Kyu Han, Ryan McDonald, Kilian~Q
  Weinberger, and Yoav Artzi.
\newblock Wav2seq: Pre-training speech-to-text encoder-decoder models using
  pseudo languages.
\newblock {\em CoRR}, 2022.

\bibitem[\protect\citeauthoryear{Xu \bgroup \em et al.\egroup
  }{2021}]{Xu_ACL2021}
Chen Xu, Bojie Hu, Yanyang Li, Yuhao Zhang, Shen Huang, Qi~Ju, Tong Xiao, and
  Jingbo Zhu.
\newblock Stacked acoustic-and-textual encoding: Integrating the pre-trained
  models into speech translation encoders.
\newblock In {\em ACL}, 2021.

\bibitem[\protect\citeauthoryear{Ye \bgroup \em et al.\egroup
  }{2021}]{Ye_interspeech2021}
Rong Ye, Mingxuan Wang, and Lei Li.
\newblock End-to-end speech translation via cross-modal progressive training.
\newblock In {\em InterSpeech}, 2021.

\bibitem[\protect\citeauthoryear{Ye \bgroup \em et al.\egroup
  }{2022}]{Ye_NAACL2022}
Rong Ye, Mingxuan Wang, and Lei Li.
\newblock Cross-modal contrastive learning for speech translation.
\newblock In {\em NAACL}, 2022.

\bibitem[\protect\citeauthoryear{Zheng \bgroup \em et al.\egroup
  }{2021}]{Zheng_ICML2021}
Renjie Zheng, Junkun Chen, Mingbo Ma, and Liang Huang.
\newblock Fused acoustic and text encoding for multimodal bilingual pretraining
  and speech translation.
\newblock In {\em ICML}, 2021.

\end{thebibliography}

\end{document}